\documentclass[a4paper,fleqn]{cas-sc}

\usepackage[authoryear,longnamesfirst]{natbib}
\usepackage[T1]{fontenc}
\usepackage[latin9]{inputenc}
\usepackage{verbatim}
\usepackage{amsmath}
\usepackage{graphicx}
\usepackage{algorithm,algpseudocode}
\usepackage[caption=false]{subfig}
\usepackage{babel}

\def\tsc#1{\csdef{#1}{\textsc{\lowercase{#1}}\xspace}}
\tsc{WGM}
\tsc{QE}
\tsc{EP}
\tsc{PMS}
\tsc{BEC}
\tsc{DE}

\begin{document}
\let\WriteBookmarks\relax
\def\floatpagepagefraction{1}
\def\textpagefraction{.001}
\shorttitle{Unsupervised representations of visual patterns}
\shortauthors{Luis Sa-Couto et~al.}

\title [mode = title]{Using brain inspired principles to unsupervisedly learn good representations
for visual pattern recognition}                      
\tnotemark[1]

\tnotetext[1]{This work was supported by national funds through Funda\c{c}\~{a}o para a Ci\^{e}ncia e Tecnologia (FCT) with reference UID/CEC/50021/2020 and through doctoral grant SFRH/BD/144560/2019 awarded to the first author. The funders had no role in study design, data collection and analysis, decision to publish, or preparation of the manuscript. The authors declare no conflicts of interest.}


\author[1]{Luis Sa-Couto}
\cormark[1]
\fnmark[1]
\ead{luis.sa.couto@tecnico.ulisboa.pt}


\address[1]{Department of Computer Science and Engineering, INESC-ID  \&  Instituto Superior T\'ecnico University of Lisbon, Av. Prof. Dr. An\'ibal Cavaco Silva, 2744-016 Porto Salvo, Portugal}

\author%
[1]
{Andreas Wichert}
\fnmark[1]
\ead{andreas.wichert@tenico.ulisboa.pt}

\cortext[cor1]{Corresponding author}



\begin{abstract}
Although deep learning has solved difficult problems in visual pattern
recognition, it is mostly successful in tasks where there are lots
of labeled training data available. Furthermore, the global back-propagation
based training rule and the amount of employed layers represents a
departure from biological inspiration. The brain is able to perform
most of these tasks in a very general way from limited to no labeled
data. For these reasons it is still a key research question to look
into computational principles in the brain that can help guide models
to unsupervisedly learn good representations which can then be used
to perform tasks like classification. In this work we explore some
of these principles to generate such representations for the MNIST
data set. We compare the obtained results with similar recent works
and verify extremely competitive results.
\end{abstract}




\maketitle

\section{Introduction}

Deep Learning is now regarded as the main paradigm to solve most learning
problems in natural tasks like vision \cite{Goodfellow2016}. And
rightfully so since the results on many tasks have been astounding
\cite{AlexNet}. Yet, this is mainly the case when large sets of labeled
data are available. Although there was some inspiration at the beginning
\cite{McCulloch1943,Rosenblatt1958,Rumelhart1986,Fukushima1980,Lecun1995},
the Deep Learning approach has mainly departed from brain related
principles. However, the brain is the best natural example of a learning
system that can perform extremely difficult tasks in a mostly unsupervised
manner \cite{CompNeuroBook}. For that reason, it is still a relevant
and active area of research to look into some established computational
brain principles to help guide the development of different models
that can unsupervisedly learn useful representations to solve difficult
tasks \cite{BAMI2016,itheory2016,BioDeepLearning2019,UnsupervisedRepresentations,Dileep2017,Sa-Couto2019,Dileep2020}.
Due to the lack of a well established theoretical understanding of
the brain one can be overwhelmed by the tons of separate pieces of
information related to it. To this end, it is helpful to recall Marr's
three levels \cite{Marr} and abstract away much of the complexity
that can come from specific neural implementations and networks. In
this work, much like in \cite{Dileep2020}, we try to identify some
key computational principles and constraints which are established
about how the brain processes visual data and implement them in a
model such that it is able to build useful unsupervised representations
of simple images. Therefore, we structure the paper around four key
principles:
\begin{enumerate}
\item What-where separation of information: since early experiments on the
early layers of the visual cortex \cite{Hubel1962,Hubel1968} that
there is evidence for structures that are specific to identify particular
stimuli (what) and others that are modelers of position (where).
\item Locality of learning: due to physical limitations, study on how the
brain learns points to the locality of the updates \cite{Hebb1949,HebbRules,AttractorBook,CompNeuroBook,BioDeepLearning2019}.
\item Vision is a temporal task: although most machine learning approaches
to image data use the pixel matrix as a fixed input, it is well established
that the brain processes vision through time with the eyes changing
position \cite{Saccades,BAMI2016}.
\item The central area of the retina (i.e. the fovea) has a detailed view
of the image, while outer areas have only a blurred view \cite{Fovea}.
\end{enumerate}
We use section \ref{sec:principles} to further detail these principles
and how they have appeared in the literature. Then, we use section
\ref{sec:model} to provide a detailed description of how our proposed
model aims to implement the principles. Section \ref{sec:experiments}
not only presents an empirical view of the model and on how to choose
the hyper-parameters, but also applies the typical experiment applied
in similar works \cite{UnsupervisedRepresentations,BioDeepLearning2019}
using the MNIST \cite{MNIST} data set. Finally, we conclude the paper
in section \ref{sec:conclusion} with some take aways and outlining
some possible paths forward.

\section{The principles in the literature\label{sec:principles}}

Hubel and Wiesel's experiments on the early stages of the visual cortex
found two specific types of cells named simple and complex \cite{Hubel1962,Hubel1968,Hubel1988,CompNeuroBook}.
Simple cells are tuned to specific stimuli like oriented lines. One
might say that they are concerned about modeling what the input is.
On the other hand, complex cells seem to react to the same stimuli
but allowing for shifts in position. These cells can be seen as modeling
the positional information of where the stimulus occurs. This idea
of first modeling the ``what'' of the stimulus and then modeling
the ``where'' of that same stimulus inspired many learning architectures
for visual pattern recognition. The seminal Neocognitron \cite{Fukushima1980}
tried to implement Hubel and Wiesel's discoveries almost directly.
This model then led to a few generalizations \cite{Cardoso2010} and
improvements \cite{Fukushima2003} to increase performance. In a very
biologically inspired implementation of the same principles the well-known
HMAX approach was proposed \cite{HMAX1} and succeeded on several
tasks \cite{HMAX2}, which led to increased interest in it with several
developments and new versions \cite{HMAX3,itheory2016}. A parallel
path, took the Neocognitron and departed from biological realism to
achieve the powerful engineering tool of convolution networks \cite{Lecun1995,LeCun1998}.
With increasing computational power and many advances in the Deep
Learning approach to train large networks, convolution networks became
the most successful member of this family at solving hard tasks \cite{Goodfellow2016,AlexNet}.
With all of that it seems that there may be something helpful about
this principle and, for that reason, we will use it to guide the development
of the proposed model.

Although back-propagation of gradients is the key approach behind
learning successful deep networks, the biological plausibility of
such dependences is questionable \cite{CompNeuroBook}. In fact, neuroscience
literature seems to point to local learning rules that are inspired
by Hebb's hypotheses \cite{Hebb1949,HebbRules}. For that reason,
a lot of research effort has been put into developing alternative
learning schemes that work layer-wise \cite{Attractor,UnsupervisedRepresentations,BioDeepLearning2019}.
With that, we will build a model in a way that all learning requires
only local information from consecutive layers.

Most machine learning approaches to image data use the pixel matrix
as a fixed input \cite{Lecun1995,AlexNet}. However, it is well established
that the brain processes vision through time with the eyes changing
position. For instance, there is extremely interesting literature
around the role that saccadic eye movements play in recognition \cite{Saccades}.
With that, we will follow the lines of related research \cite{BAMI2016}
and include a time component in our processing.

At a given moment in time, the image that is projected in the retina
is described at different resolutions. More specifically, the central
area of the retina (i.e. the fovea) has a detailed view of the image,
while outer areas have only a blurred view \cite{Fovea}. It is often
posited that the central area can focus on detail while the outer
areas offer the context where the detail is inserted. In our opinion,
this view, combined with further neuroscientific evidence, suggests
that the brain may use outer retinal information to have a notion
about the relative position of the detail in the object. This led
us to posit an object-dependent frame in previous work on developing
sparse codes for an associative memory task \cite{Sa-Couto2019} and
for classification \cite{Sa-Couto2020}. In this paper we will further
detail this idea and include this kind of processing in our model.

\section{Putting the principles in a model\label{sec:model}}

\begin{figure}
\begin{centering}
\includegraphics[scale=0.5]{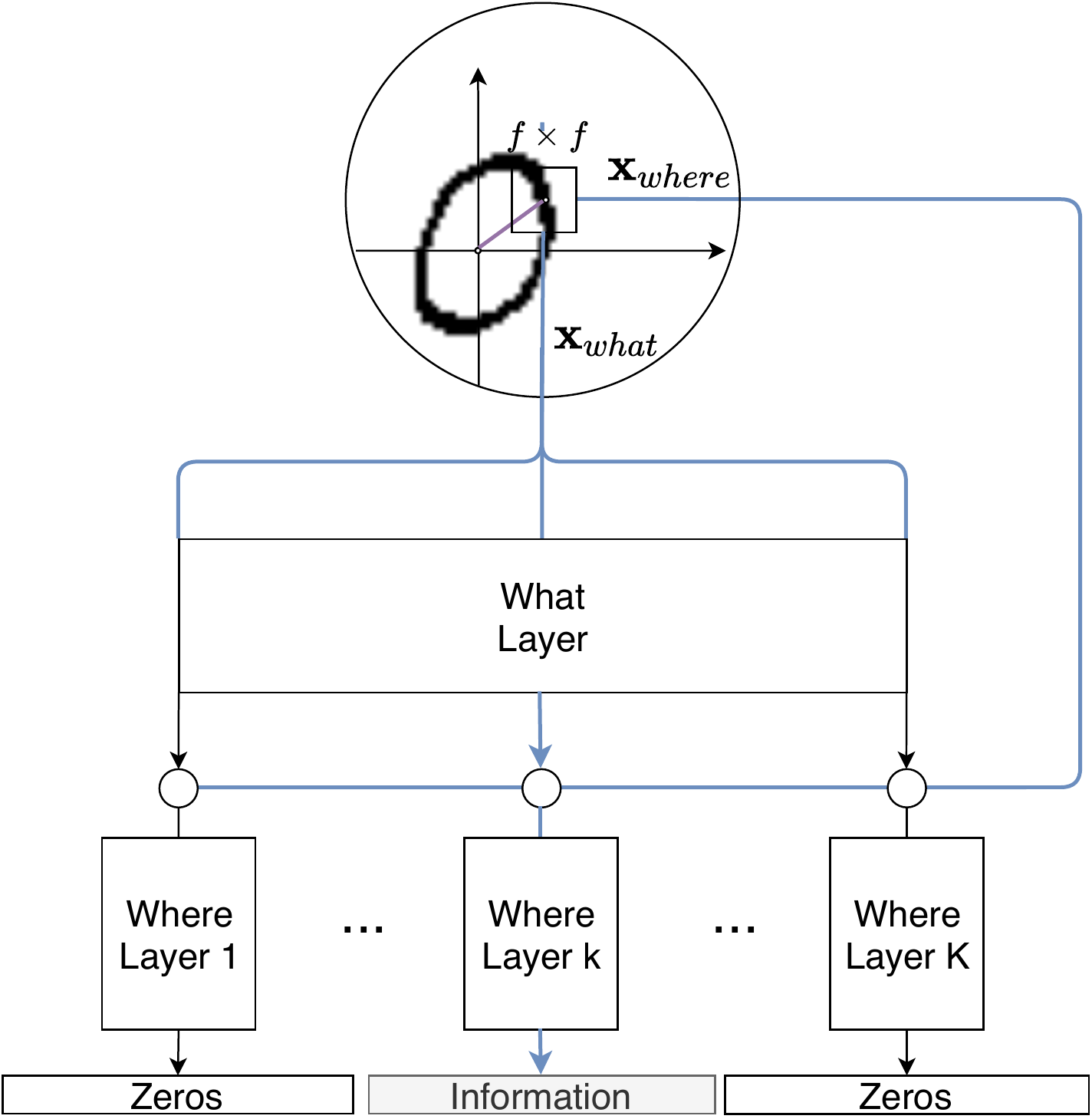}
\par\end{centering}
\caption{At a given instant, an $f\times f$ window $\mathbf{x}_{what}$ of
the image is taken as input to the what layer. The what layer then
identifies the where layer that corresponds to this particular input.
That where layer receives as input the window's object-dependent position
$\mathbf{x}_{where}$ and based on that provides an encoding of the
information. Looking at the output as a vector, we can say that all
where layers that were not chosen by the what layer provide an output
of zeros while the chosen layer provides an encoding for the information.
\label{fig:Module_view}}
\end{figure}

Following the aforementioned third and fourth principles, we define
an $f\times f$ sized window that represents the high detailed region
of the model's view of the image at a particular time step. We can
look at the content of this region as a vector $\mathbf{x}_{what}$.
Furthermore, we use the outer area's information to find the position
of this content window in an object-dependent coordinate system (we
will see later how). We can look at this position as a vector $\mathbf{x}_{where}$.
This logic is illustrated in figure \ref{fig:Module_view}.

With that, at each moment in time we get two pieces of information.
Following the what-where principle, and looking at figure \ref{fig:Module_view}
for an illustration, the model starts by processing $\mathbf{x}_{what}$.
To that end, there is a what layer, which will be detailed in subsection
\ref{subsec:What}. This layer performs recognition of the content
and directs processing to a where layer that is specific for that
type of stimulus. As the name indicates, the where layer models the
positional information $\mathbf{x}_{where}$ and will also be detailed
later (in subsection \ref{subsec:Where}).

From this abstract view of the model, we can see that, for each time
step, we will generate a vector that encodes information about what
was seen and where it was seen relative to the global object. To get
at a single, final representation for the whole object we will need
to combine all of these vectors into one. We will discuss our approach
to this issue in subsection \ref{subsec:Pooling}.

\subsection{What\label{subsec:What}}

In this subsection we open the abstract ``what layer'' box in figure
\ref{fig:Module_view}. Besides describing its operation we also detail
which hyper-parameters are involved and how to learn the trainable
parameters.

\subsubsection{Feature Mapping}

The what layer implements the winner takes all approach to feature
mapping \cite{Cardoso2010}. Each of $K$ units is tuned to recognize
a given preferred pattern $\mathbf{w}_{k},k=1,\ldots,K$ (like a corner
or an oriented line). Given an input, each unit measures a cosine
similarity between its preferred pattern and that input \cite{Sa-Couto2019}.
The usage of this measure can be viewed as applying weight normalization
in a typical dot product based layer. Such normalization is also biologically
plausible since synaptic strength cannot grow unbounded \cite{AttractorBook,CompNeuroBook}.
The units then compete and the most similar one wins firing a $1$
while the others output $0$. The usage of an absolute minimum threshold
$T$ ensures that there is not always a winner. For inputs that do
not resemble any of the preferred patterns, all units will be silent.

To implement this reasoning, we write the net input to unit $k$ with
equation \ref{eq:cosine}.

\begin{equation}
net_{k}=\frac{\mathbf{x}_{what}^{T}\mathbf{w}_{k}}{\left\Vert \mathbf{x}_{what}^{T}\right\Vert \left\Vert \mathbf{w}_{k}\right\Vert }\label{eq:cosine}
\end{equation}

To define the binary activations of each unit we use the well-known,
right continuous, Heaviside step activation function given in equation
\ref{eq:heaviside}.

\begin{equation}
H^{^{+}}\left(x\right)=\begin{cases}
1 & x\geq0\\
0 & x<0
\end{cases}\label{eq:heaviside}
\end{equation}

Unit $k$'s output, written $what_{k}$, is the result of the competition
between the layer's units and it can be written with equation \ref{eq:what_output}.

\begin{equation}
what_{k}=H^{+}\left(net_{k}-\max\left(T,\max_{l\in\left\{ 1,\cdots,K\right\} }net_{l}\right)\right)\label{eq:what_output}
\end{equation}

\begin{figure}
\begin{centering}
\includegraphics[scale=0.5]{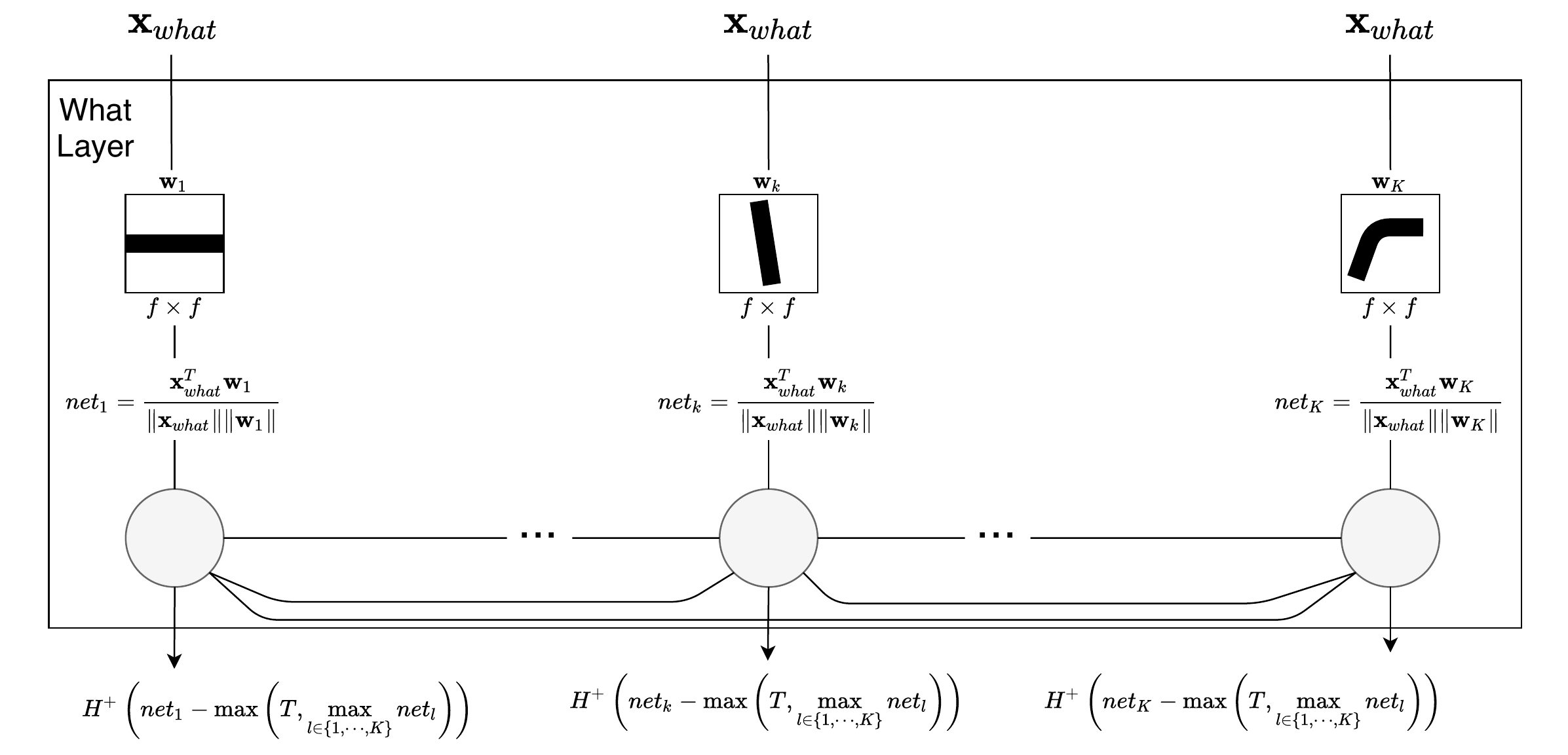}
\par\end{centering}
\caption{The what layer implements the winner takes all approach to feature
mapping. Each of $K$ units is tuned to recognize a given preferred
pattern $\mathbf{w}$ (like a corner or an oriented line). Given an
input, each unit measures a cosine similarity between its preferred
pattern and that input. The units then compete and the most similar
one wins firing a $1$ while the others output $0$. The usage of
an absolute minimum threshold $T$ ensures that there is not always
a winner. For inputs that do not resemble any of the preferred patterns,
all units will be silent. \label{fig:What_view}}
\end{figure}

Figure \ref{fig:What_view} provides an illustration of information
processing in the what layer.

\subsubsection{Competitive Learning}

Now that we have described the operation we are left with the learning
problem: how to learn the preferred patterns $\mathbf{w}_{k}$? To
this end we employ the typical competitive learning approach \cite{CompetetiveLearning1,AttractorBook,Haykin2008}
where for a given input, the winner unit gets its weights updated.
One can also look at this learning approach as a variant of k-means
clustering \cite{KMeans1} applied in a stochastic manner to minibatches
\cite{MiniBatchKmeans}. All in all, we can describe the learning
procedure with the rule in equation \ref{eq:competitive} where $\eta_{k}$
is the learning rate.

\begin{equation}
\mathbf{w}_{k}=\mathbf{w}_{k}+what_{k}\eta_{k}\left(\mathbf{x}_{p}-\mathbf{w}_{k}\right)\label{eq:competitive}
\end{equation}

Besides adjusting the learnable parameters, $T$, $K$ and $f$ play
the role of hyper-parameters and have to be chosen based on the task
at hand. In section \ref{sec:experiments} we will discuss how we
chose them for specific experiments.

\subsection{Where\label{subsec:Where}}

In this subsection we open one of the abstract ``where layer'' boxes
in figure \ref{fig:Module_view}. Besides describing its operation
we also detail which hyper-parameters are involved an how to learn
the trainable parameters. In general, we can describe the operation
of this layer as using a Gaussian Mixture Model \cite{Bishop2006,MurphyBook}
of positions in the object-dependent space.

\subsubsection{Object dependent frame}

The first important issue comes in the definition of the object-dependent
frame. Assuming that $\mathbf{x}_{img}$ corresponds to the window's
position in the pixel matrix. To transform this position to the new
coordinate system we need a center and a radius. To compute them,
we need to know which positions belong to the object. In a more realistic
scenario, one could use stereopsis and depth combined with color to
achieve this (or even a segmentation model). However, since this is
not the main point of this work, we abstract away this complexity
and use the same strategy as in \cite{Sa-Couto2019,Sa-Couto2020}
and state that a position belongs to the object if its what layer
activation is nonzero. Using that strategy, we can use equation \ref{eq:center}
to compute the center

\begin{equation}
C=\frac{\sum_{\mathbf{x}_{img}}\sum_{k=1}^{K}\mathbf{x}_{img}what_{k}\left(\mathbf{x}_{img}\right)}{\sum_{\mathbf{x}_{img}}\sum_{k=1}^{K}what_{k}\left(\mathbf{x}_{img}\right)}\label{eq:center}
\end{equation}
and equation \ref{eq:radius} to compute the radius.

\begin{equation}
R=\max_{\mathbf{x}_{img},k\in\left\{ 1,\ldots,K\right\} }\left\{ what_{k}\left(\mathbf{x}_{img}\right)\left\Vert \mathbf{x}_{img}-C\right\Vert _{2}\right\} \label{eq:radius}
\end{equation}
With that, we can map $\mathbf{x}_{img}$ to the intended $\mathbf{x}_{where}$
using equation \ref{eq:object_frame}.

\begin{equation}
\mathbf{x}_{where}=\frac{\mathbf{x}_{img}-C}{R}\label{eq:object_frame}
\end{equation}

\subsubsection{Positional Mapping}

To implement the Gaussian mixture, each unit $l\in\left\{ 1,\ldots,C_{k}\right\} $
in where layer $k$ is parameterized by a weight $\pi_{l}^{k}$, a
center $\mu_{l}^{k}$ and a covariance $\mathbf{\Sigma}_{l}^{k}$.
The net input to a unit is the unnormalized Gaussian probability assigned
by that unit to that particular position. This is expressed in equation
\ref{eq:gmm}.

\begin{equation}
net_{l}^{k}=\pi_{l}^{k}\mathcal{N}\left(\mathbf{x}_{where}\mid\mathbf{\mu}_{l}^{k},\mathbf{\Sigma}_{l}^{k}\right)\label{eq:gmm}
\end{equation}
One can interpret the mean and covariance as describing a receptive
field over positions.

The final output of each unit is also the product of competition between
lateral units as is written in equation \ref{eq:where_output}. This
is basically the normalization of the probabilities.

\begin{equation}
where_{l}^{k}=\frac{net_{l}}{\sum_{i=1}^{C_{k}}net_{i}}\label{eq:where_output}
\end{equation}

\begin{figure}
\begin{centering}
\includegraphics[scale=0.5]{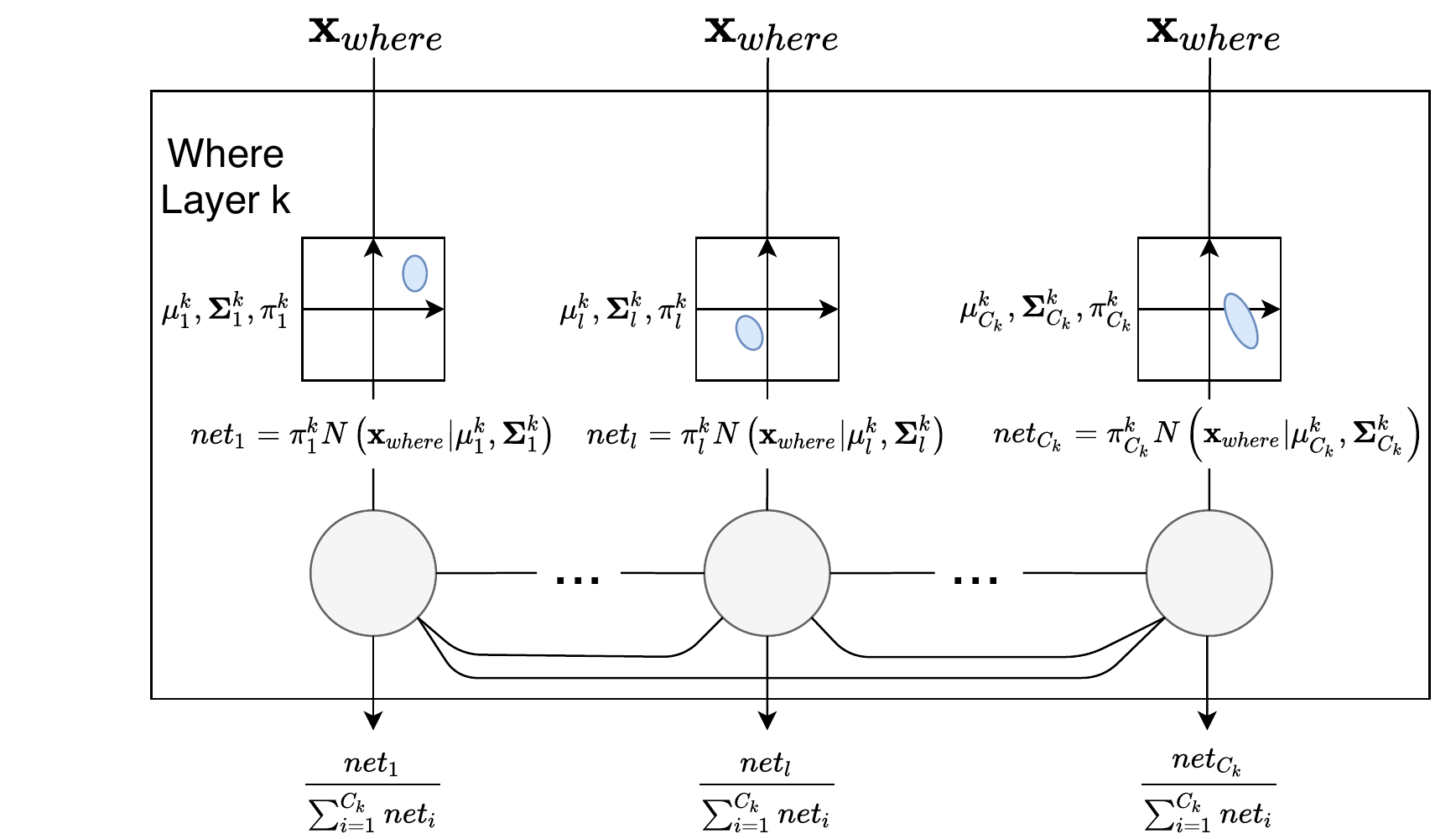}
\par\end{centering}
\caption{Where layer $k$ has $C_{k}$ units that describe a Gaussian mixture
over the space of object-dependent positions of occurrences of a given
pattern. Each unit has a weight, a mean vector and a covariance that
together describe the area of frequently occurring positions. The
final output is a soft competition to decide which component generated
the observed position. \label{fig:Where_view}}
\end{figure}

With this description, we see that, since each unit represents a component,
the layer's operations is a competition to see from which component
the position was generated. Figure \ref{fig:Where_view} provides
an illustration of the layer's operation. 

\subsubsection{Expectation Maximization Learning}

To learn the parameters we apply the typical approach to learn a Gaussian
Mixture. More specifically, we use expectation-maximization \cite{Bishop2006}.
For a set of $P$ examples, the expectation step corresponds to computing
the layer's output for each one. The maximization step is just the
maximum likelihood estimation of each parameter using equations \ref{eq:em_mu},
\ref{eq:em_sigma} and \ref{eq:em_pi}.

\begin{equation}
\mu_{l}^{k}=\frac{1}{\sum_{p=1}^{P}where_{pl}^{k}}\sum_{p=1}^{P}where_{pl}^{k}\mathbf{x}_{p}\label{eq:em_mu}
\end{equation}

\begin{equation}
\Sigma_{l}^{k}=\frac{1}{\sum_{p=1}^{P}where_{pl}^{k}}\sum_{p=1}^{P}where_{pl}^{k}\left(\mathbf{x}_{p}-\mathbf{\mu}_{l}^{k}\right)\left(\mathbf{x}_{p}-\mathbf{\mu}_{l}^{k}\right)^{T}\label{eq:em_sigma}
\end{equation}

\begin{equation}
\pi_{l}^{k}=\frac{\sum_{p=1}^{P}where_{pl}^{k}}{\sum_{p=1}^{P}\sum_{l'=1}^{C_{k}}where_{pl'}^{k}}\label{eq:em_pi}
\end{equation}

Besides the learnable parameters, it is a key architectural feature
to decide how many units each where layer uses. In section \ref{sec:experiments}
we will use a heuristic way of making this decision for a specific
task.

\begin{figure}
\begin{centering}
\includegraphics[scale=0.3]{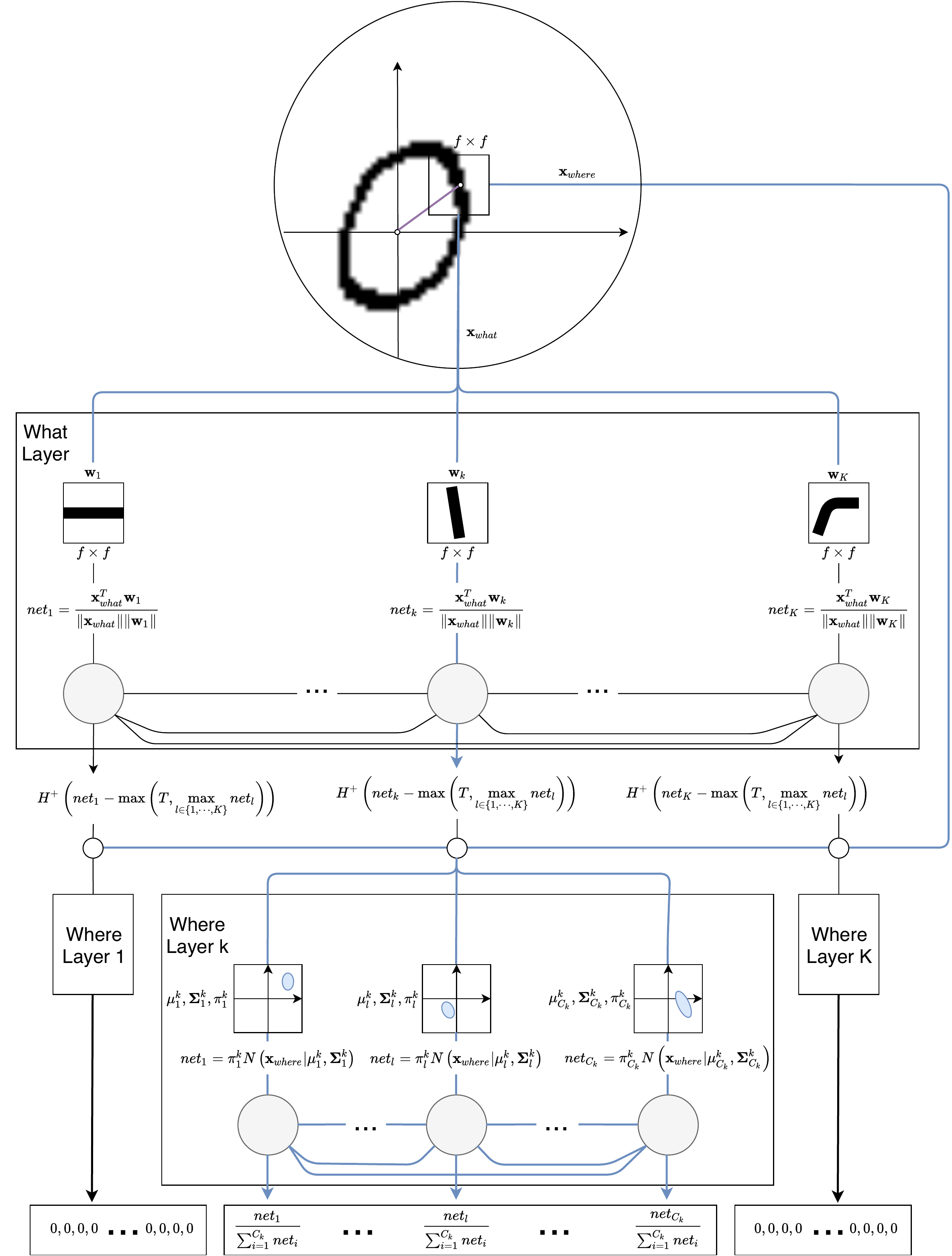}
\par\end{centering}
\caption{A detailed view of the processing that occurs for a given time step.
\label{fig:Detailed_view}}
\end{figure}

Now that we have lifted the lids from the abstracted view in figure
\ref{fig:Module_view} we can put it all together in a detailed view
(see figure \ref{fig:Detailed_view}) of the processing at a given
time step.

\subsection{Pooling Sparse Views to get a global view\label{subsec:Pooling}}

\begin{figure}
\begin{centering}
\includegraphics[scale=0.3]{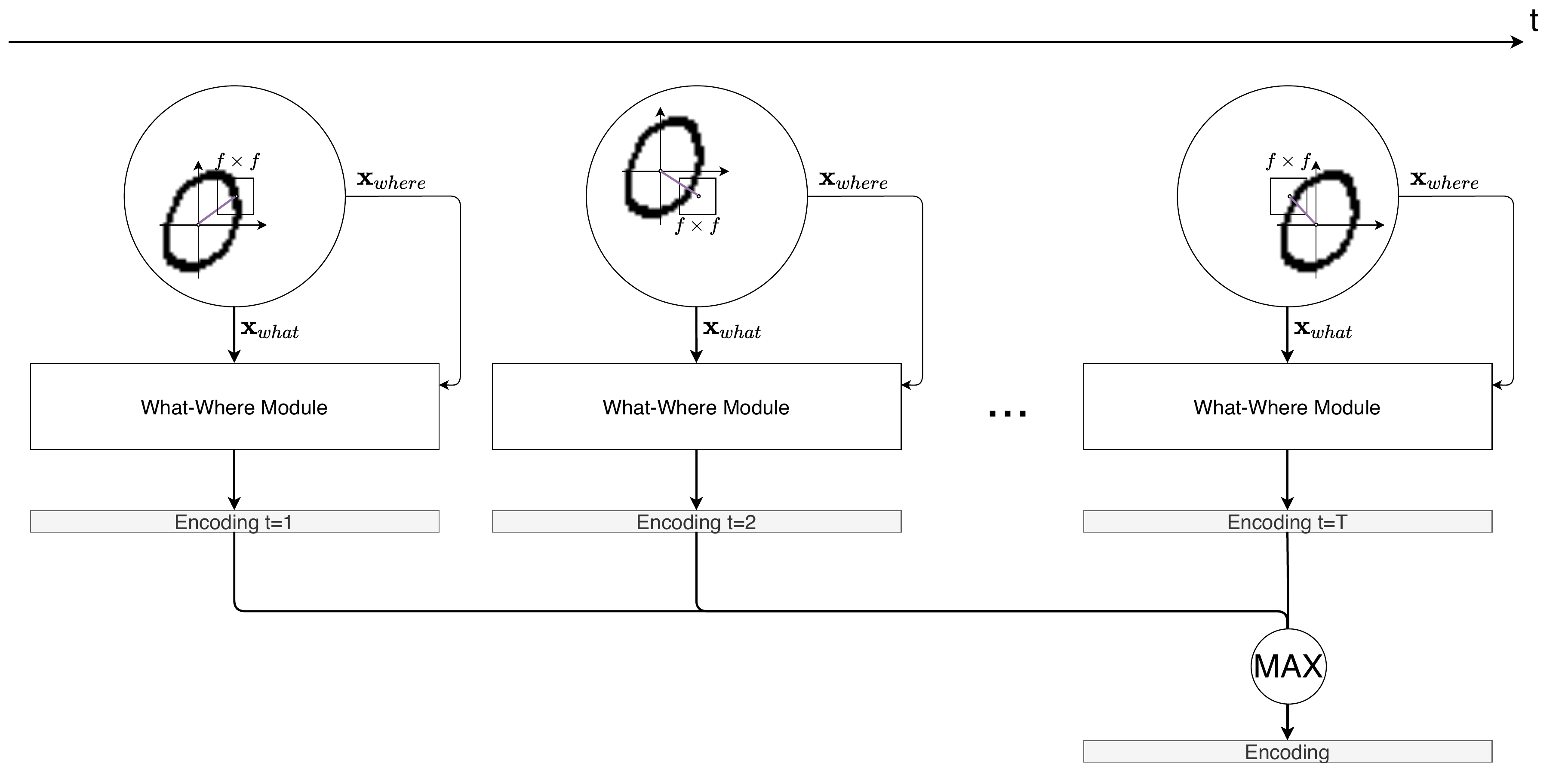}
\par\end{centering}
\caption{The what-where module is applied at every time step producing, for
each, a description. To get a final vector representation, we apply
element-wise max pooling. This yields a probabilistic presence map
that states wether a given feature has appeared at a given position.
\label{fig:Temporal_view}}
\end{figure}

As we have stated before, with a vector description for each time
step we need a way to combine all descriptions into a final one. To
get a probabilistic presence map telling wether a given feature has
appeared at a given position or not, it makes sense to use element-wise
max pooling as described by equation \ref{eq:time_max}.

\begin{equation}
o_{l}^{k}=\max_{t\in\left\{ 1,\ldots,T\right\} }where_{l}^{k}\left(t\right)\label{eq:time_max}
\end{equation}
This idea is illustrated in figure \ref{fig:Temporal_view}.

Many strategies may be implemented on how many time steps should be
used and to which positions the model should jump at each one. Further
research into saccadic movements may be required to develop the best
possible approach. Since this is not the main purpose of this work,
in the next sections, all experiments have a time step for each possible
image position.

\section{Experiments\label{sec:experiments}}

The MNIST data set \cite{MNIST} contains a training set of $60000$
images of $28\times28$ pixels with handwritten digits. It has also
a fixed test set with $10000$ examples where models can be evaluated
comparably. In this section, we make use of this data to show how
the model works in practice. We start by detailing the results of
the unsupervised layer in the what and where layers. After that, we
use the unsupervised representations of the data to learn a linear
classifier and evaluate the test set accuracy. This is the typical
approach used in comparable works \cite{BioDeepLearning2019,UnsupervisedRepresentations}
to evaluate the quality of the generated representations.

\subsection{What Stage Features}

As was mentioned beforehand, there are three hyper-parameters that
play a role at this stage:
\begin{enumerate}
\item $T$: the minimum similarity the winner unit needs to get for a winner
to exist.
\item $f$: the size of the side of the detailed view window in pixels.
\item $K$: the number of units in the layer.
\end{enumerate}
There is a connection between $T$ and $K$. As the number of units
increases, the probability that a given input will be very dissimilar
to all of the preferred patterns in the layer decreases. So, provided
that $K$ is large enough, in several experiments we observe that
the model is relatively robust to the choice of $T$. As long as the
value is not too high, in which case too much information is lost,
the model works similarly for most values. Basically, the problem
becomes the choice of $K$. Fortunately, we can use the mapping between
competitive learning and $K$-means clustering to exploit the vast
amount of literature that exists on choosing the number of clusters
\cite{KMeansPlasticity,Kselection}. Although many techniques become
available, we can also just choose this parameter through random search.
By doing so, we again verify a relative robustness to the choice from
a large enough value upward.

We also need to choose $f$. The idea is to capture local features
of the visual object so this value must not be too large. Furthermore,
a large value can cause the curse of dimensionality to ruin the learning
process \cite{Bishop2006}. For that reason, we have tried a few different
small values (i.e. $f=3,5,7,9$) and found $f=5$ to work best for
this data set.

\begin{figure}
\begin{centering}
\includegraphics[scale=0.5]{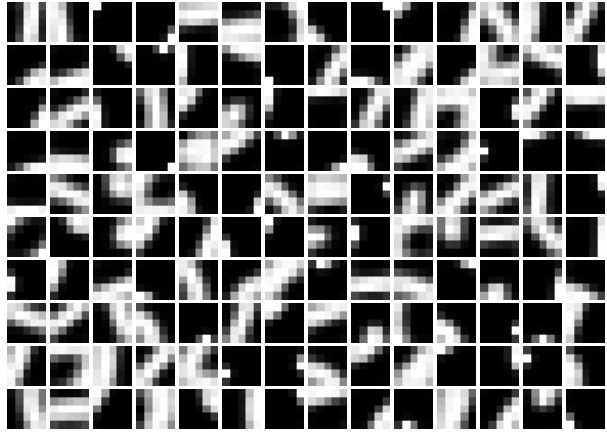}
\par\end{centering}
\caption{The what layer features learned through competitive learning for the
top performer model in subsection \ref{subsec:Classification}. The
what layer hyper-parameters are, in this case, $K=140,f=5,T=0.7$.
\label{fig:What-features}}
\end{figure}

As was mentioned before, in subsection \ref{subsec:Classification}
we will apply the typical classification experiment to the model.
In figure \ref{fig:What-features} we plot the result of competitive
learning for the top performer in that experiment. As expected, the
learned features are mostly oriented lines a corners.

\subsection{Where Stage Positions}

For a what layer with $K$ units, the model will have $K$ where layers.
For each of these layers we will need to choose the number of units.
For that reason, unlike in the previous section, a random search would
be extremely time consuming. Fortunately, we can look at the learning
problem as a density estimation with a Gaussian mixture model and,
by doing so, we can use one of the many techniques that have been
proposed to choose the number of Gaussian components. In these experiments,
we apply the bayesian information criterion (BIC) \cite{MurphyBook}.

To understand this criterion, let us assume we are choosing the number
of units in where layer $k$. Assuming $L$ is the data's log-likelihood
under a given mixture model and $P$ is again the number of examples,
the BIC value is given by equation \ref{eq:bic}

\begin{equation}
BIC\left(C_{k}\right)=2\log L-\Omega_{gmm}\left(C_{k}\right)\log P\label{eq:bic}
\end{equation}
where $\Omega_{gmm}\left(C_{k}\right)$ is the number of parameters
in a GMM with $C_{k}$ Gaussians. In our specific case, this will
always be a two-dimensional problem where we will need:
\begin{itemize}
\item $C_{k}$ parameters to represent the $\pi_{l}^{k}$ values with $l=1,\ldots,C_{k}$.
\item $2C_{k}$ parameters to represent the mean vector coordinates for
each component.
\item $3C_{k}$ parameters to represent, for each component, the variance
of each dimension and the covariance between dimensions.
\end{itemize}
With that in mind, the number of parameters will be given by equation
\ref{eq:gmm_params}.

\begin{equation}
\Omega_{gmm}\left(C_{k}\right)=\Omega\left(\mathbf{\pi}^{k}\right)+\Omega\left(\mathbf{\Sigma}^{k}\right)+\Omega\left(\mathbf{\mu}^{k}\right)=C_{k}+3C_{k}+2C_{k}\label{eq:gmm_params}
\end{equation}

A choice of $C_{k}$ that scores high in the BIC is going to yield
a compromise between complexity and explanatory power. That is, a
model that explains the data well with as few parameters as possible.
To make this choice, we apply something close to the elbow method
\cite{Kselection} where we progressively increase $C_{k}$ and measure
$\Delta BIC=BIC\left(C_{k}+1\right)-BIC\left(C_{k}\right)$. When
the improvement is below a given threshold $\Delta BIC<T_{bic}$ we
stop the search. In practice, this adds a new hyper-parameter $T_{bic}$
to the model and in the next subsection we will see results with different
choices.

\begin{figure}
\begin{centering}
\subfloat[]{\begin{centering}
\includegraphics[scale=0.5]{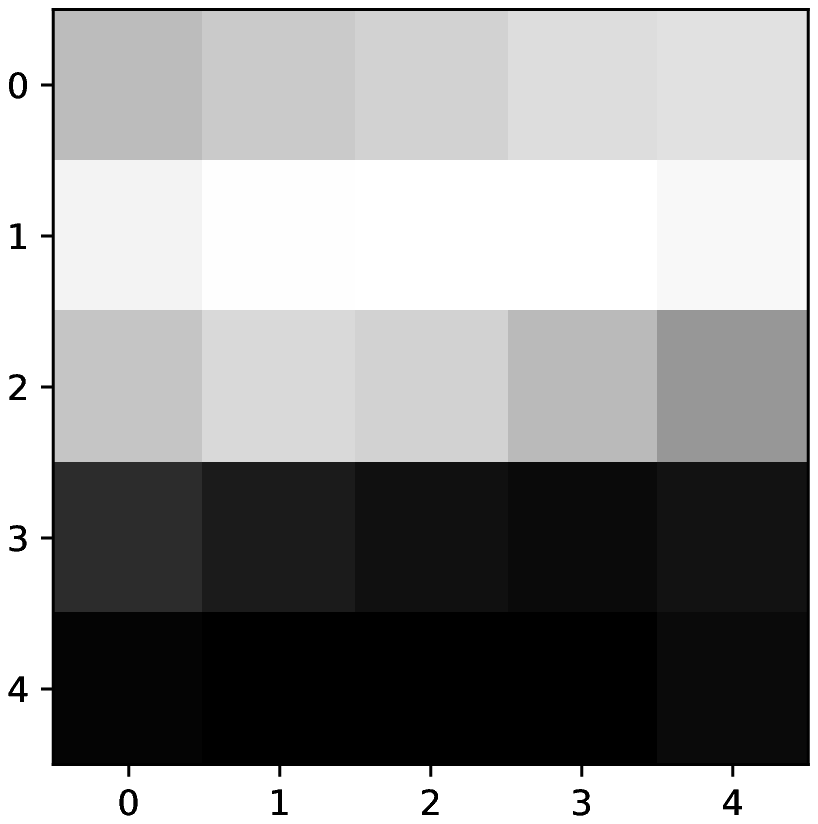}
\par\end{centering}

}\subfloat[]{\begin{centering}
\includegraphics[scale=0.5]{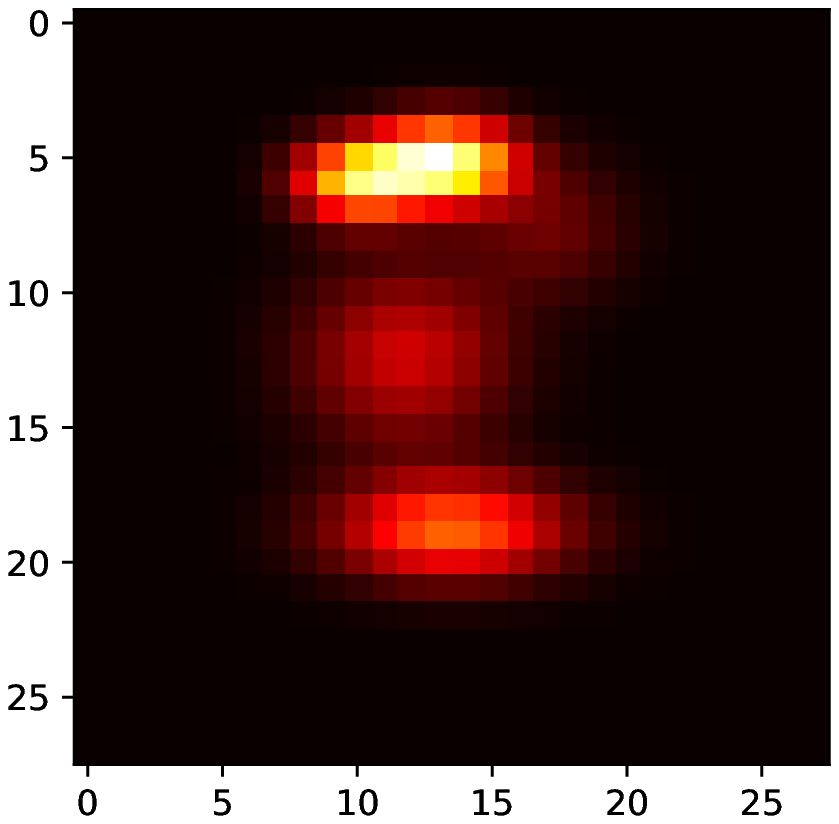}
\par\end{centering}

}\caption{The ``what'' feature presented in a) leads to a where mixture model
with probability heatmap on a). This is one of several particularly
intuitive results where, thinking about digits, an horizontal line
would appear at three possible regions: top, middle or bottom. \label{fig:where-feature}}
\par\end{centering}
\end{figure}

To get an intuition for what a where layer learns we provide figure
\ref{fig:where-feature} which shows the resulting model for a specific
what feature.

\subsection{Classification results\label{subsec:Classification}}

Like in comparable works \cite{BioDeepLearning2019,UnsupervisedRepresentations}
we use a linear classifier trained with the model's unsupervised representations
to see how good they are. In this case, we used a simple generalized
perceptron \cite{Rosenblatt1958} or, more specifically, a logistic
regression \cite{Bishop2006,Goodfellow2016} output layer. Some results
for different hyper-parameter choices are given in table \ref{tab:accuracy}.
We see that the model performs well which tells us that the learned
representations are capturing some real information about the underlying
patterns.
\begin{center}
\begin{table}
\begin{centering}
\begin{tabular}{|c|c|c|c|}
\hline 
$T$ & $K$ & $T_{bic}$ & Test set Accuracy\tabularnewline
\hline 
\hline 
0.7 & 60 & 10 & 99.12\tabularnewline
\hline 
0.6 & 130 & 10 & 99.18\tabularnewline
\hline 
0.5 & 80 & 1 & 99.18\tabularnewline
\hline 
0.7 & 140 & 5 & 99.24\tabularnewline
\hline 
\end{tabular}
\par\end{centering}
\caption{Some representative classification results on the $10000$ MNIST test
set images for different hyper-parameters. In these type of model
it is usually said that success is scoring above $98\%$ \cite{BioDeepLearning2019}
\label{tab:accuracy}}

\end{table}
\par\end{center}

\section{Conclusion\label{sec:conclusion}}

In this work we leveraged some well established principles about the
brain to unsupervisedly learn effective representations for simple
visual data. We started by identifying the four key principles and
showing how they appear throughout the literature. Afterward, we went
into model details showing how they could be implemented. We then
discussed the model's hyper-parameters and principled ways to choose
values for them. Finally, we employed the typical approach of using
a linear classifier to evaluate the quality of the generated representations
for the MNIST data set and verified that we get extremely competitive
results. We recognize a few limitations that can lead to next steps:
tougher images would require a few extra details, namely we could
use color, texture or depth from stereopsis to build the object-dependent
frame; moving downward in Marr's levels \cite{Marr} we could build
a more neurally detailed implementation of the model. However, even
with such possible ways forward, we have shown that these principles
are useful and lead to interesting results. Furthermore, the model
has a clear interpretation of modeling content and position and its
architectural parameters can be chosen in a principled manner.

\section{Bibliography}

\printcredits

\bibliographystyle{cas-model2-names}

\bibliography{MyBib}

\end{document}